\documentclass[3p,times,twocolumn]{elsarticle}
\usepackage{ecrc}

\volume{00}
\firstpage{1}
\journalname{Procedia Computer Science}
\runauth{}
\jid{procs}
\jnltitlelogo{Procedia Computer Science}

\usepackage{amssymb}
\usepackage[figuresright]{rotating}
\usepackage{natbib}
\usepackage{tikz}
\usepackage{amsmath,amssymb}
\usepackage[toc,page]{appendix}
\usepackage{pgfplotstable}
\pgfplotsset{compat=1.17}
\usepackage{booktabs}
\usepackage{subcaption}
\usepackage{xcolor}
\usepackage{enumitem} 
\usepackage{url}
\usepackage[justification=centering]{caption}
\usepackage{hyperref}
\usepackage{bbm}
\usepackage{setspace}
\usepackage{etoolbox}
\usepackage{amsmath}
\newtheorem{proposition}{Proposition}
\newtheorem{proof}{Proof of Proposition}

\usepackage{mathabx}
\usepackage{cleveref}

\AtBeginEnvironment{quote}{\par\singlespacing\small}

\newcommand{\splitatcommas}[1]{%
  \begingroup
  \begingroup\lccode`~=`, \lowercase{\endgroup
    \edef~{\mathchar\the\mathcode`, \penalty0 \noexpand\hspace{0pt plus 1em}}%
  }\mathcode`,="8000 #1%
  \endgroup
}

\begin{document}

\begin{frontmatter}

    \title{
        Designing Time-Series Models With Hypernetworks \& Adversarial Portfolios
    }
    \author{
        Filip Staněk\fnref{cor1}
    }
    \address{
        CERGE-EI
    }
    \fntext[cor1]{
        \texttt{filip.stanek@cerge-ei.cz}\\
        CERGE-EI, a joint workplace of Charles University and the Economics Institute of the Czech Academy of Sciences, Politickych veznu 7, 111 21 Prague, Czech Republic.
    }

    \dochead{}

    \begin{abstract}
        This article describes the methods that achieved 4th and 6th place in the forecasting and investment challenges, respectively, of the M6 competition, ultimately securing the 1st place in the overall duathlon ranking.
        In the forecasting challenge, we tested a novel meta-learning model that utilizes hypernetworks to design a parametric model tailored to a specific family of forecasting tasks.
        This approach allowed us to leverage similarities observed across individual forecasting tasks while also acknowledging potential heterogeneity in their data generating processes.
        The model's training can be directly performed with backpropagation, eliminating the need for reliance on higher-order derivatives and is equivalent to a simultaneous search over the space of parametric functions and their optimal parameter values.
        The proposed model's capabilities extend beyond M6, demonstrating superiority over state-of-the-art meta-learning methods in the sinusoidal regression task and outperforming conventional parametric models on time-series from the M4 competition.
        In the investment challenge, we adjusted portfolio weights to induce greater or smaller correlation between our submission and that of other participants, depending on the current ranking, aiming to maximize the probability of achieving a good rank.
        While this portfolio strategy can increase the probability of securing a favorable rank, it paradoxically exhibits negative expected returns.
    \end{abstract}


\end{frontmatter}

\section{Introduction}

The M6 Financial Forecasting Competition \citep[see][]{makridakisM6FinancialDuathlon2022} spanned from March 2022 to February 2023 and focused on a universe of 100 assets: 50 S\&P 500 stocks and 50 international ETFs.
In the forecasting challenge, participants were tasked with predicting probabilities for each asset's next 4-week returns falling into one of five quintiles relative to other assets in the universe. 
The accuracy of these predictions was assessed using the ranked probability score (RPS) loss after the 4-week period had passed.
In the investment challenge, participants were required to submit portfolio weights for the upcoming 4-week interval. 
These portfolios were then evaluated based on risk-adjusted returns (IR).
Additionally, participants competed in a duathlon, which combined both forecasting and investment challenges. 
The duathlon ranking was computed as an arithmetic mean of participants' ranks in the forecasting and investment challenges.
This article describes the methods we employed for our submissions, which achieved 4th place in the forecasting challenge, 6th place in the investment challenge, and ultimately secured the 1st place in the duathlon.

In the forecasting challenge, we tested a novel hypernetwork meta-learning architecture capable of constructing the optimal parametric model for a given family of similar but not necessarily identical data generating processes (DGPs henceforth). 
This method, while broadly applicable, is especially well-suited for time-series forecasting, where the number of observations is typically insufficient to apply nonparametric methods on a per-series basis, but where multiple realizations of similar (but not necessarily ex-ante identical) time-series are available.

In particular, in the case of M6, this method allows us to perform a search over the space of prediction functions parameterized by some latent parameter vector specific to each asset, as opposed to finding a single prediction function for all assets, as one would do when applying a conventional nonparametric model on pooled data.
The latent parameter vector can absorb heterogeneity in DGPs across assets, hence improving performance and even allowing to leverage data on additional assets not specified by organizers without concerns that their DGPs are too dissimilar to the original M6 asset universe.

The quintile predictions, broadly speaking, encode two types of information. 
First, the relative size of predictions for the 1st and 2nd quintiles, as opposed to the 4th and 5th, provides information about whether the asset is likely to over-perform or under-perform relative to other assets.
Second, the relative size of the 1st and 5th quintiles, as opposed to the 2nd, 3rd, and 4th, generally encodes information about volatility, with assets having high predicted probabilities of both the 1st and 5th quintiles being likely to end up with more extreme returns compared to other assets.
In our case, the good performance of the model in terms of RPS is entirely driven by predicting volatility.
Consequently, our initial efforts to transform these predictions into portfolios with consistently above-normal expected IR proved unsuccessful.

In the investment challenge, due to a lack of better alternatives, we instead attempted to discretionary adjust the level of risk (as measured by the correlation between our IR and the IRs of competitors) based on our current rank within the global leaderboard.
This involved assuming more risk when the probability of achieving a favorable rank was low and less risk when a sufficiently good rank on the leaderboard had been attained.
Simulations, as well as bootstrap exercises, indicate that such an approach can indeed substantially improve the probability of securing top ranks in the leaderboard, despite paradoxically exhibiting a very poor IR in expectation.
This highlights that the task of attaining the highest expected IR and that of maximizing the probability of success in the investment challenge may not necessarily be identical, and could even be at odds with each other.

The remainder of the article is structured as follows.
Section \ref{section:forecasting_challenge} focuses on the forecasting challenge, with subsection \ref{subsectition:mtms_model} introducing the proposed meta-learning model, and subsection \ref{subsectition:mtms_application} detailing how the model was applied in the M6 competition.
Section \ref{section:investment_challenge} focuses on the investment challenge, with subsections \ref{subsection:scaling} and \ref{subsection:strategic} outlining the principles that guided the decisions, and subsection \ref{subsection:decisions} discussing the actual investment decisions made during the competition.
Section \ref{section:conclusions} concludes. 
To validate the model's effectiveness beyond the M6 competition, Appendices A and B evaluate its performance on sinusoidal regression and the M4 dataset. 
Appendices C and D contain supplementary materials and proofs, respectively.\footnote{ 
    A replication repository for this article is available at \url{https://github.com/stanek-fi/M6_article}. 
    The original repository, containing the unaltered scripts used for the submissions, is available at \url{https://github.com/stanek-fi/M6}.
}

\section{Forecasting challenge}\label{section:forecasting_challenge}
\subsection[Model]{Model\footnote{An early version of this section appeared in a pre-print \citet{stanekNoteM6Forecasting2023}.}}\label{subsectition:mtms_model}
\subsubsection{Motivation}

According to the classification by \citet{januschowskiCriteriaClassifyingForecasting2020}, time-series forecasting approaches can be broadly divided into two strains.
The conventional approach, known as \emph{local modeling}, involves selecting the most appropriate parametric model for a given family of forecasting tasks, often based on expert judgment. 
This model is then applied to each individual observed series independently. 
On the other hand, \emph{global models} consider all observed time-series jointly.
In extreme cases, this can be done via pooling, thus disregarding the information regarding which data belong to which series altogether and estimating a single global model \citep[see, e.g.,][]{montero-mansoPrinciplesAlgorithmsForecasting2021}.
However, it is often beneficial to utilize this information to help account for possible heterogeneity among the DGPs underlying the series, an approach aptly dubbed the \emph{localization of global models} \citep{godahewaEnsemblesLocalisedModels2021}. 
This is typically performed by grouping the series either with time-series clustering techniques based on time-series features \citep{bandaraForecastingTimeSeries2020} or directly according to model performance \citep{smylDataPreprocessingAugmentation2016, smylHybridMethodExponential2020} and estimating a specialized global model on each cluster. 
By adjusting the number of such clusters, one can then regulate the degree of globality/locality.

We present an alternative method that helps bridge the gap between these two extremes.
A global model, which instead of deriving a single forecasting function for all time-series, outputs a function parameterized by a latent parameter vector specific to each series, thereby acknowledging the potential heterogeneity of DGPs. 
Accounting for heterogeneity through the latent parameter space, rather than clustering, offers the advantage of equally accommodating a mixture of several different types of DGPs as well as family of continuously varying DGPs.
Alternatively, this approach can also be viewed as a data-driven alternative to manually designing a parametric model for a group of related prediction tasks, an endeavor which typically requires considerable statistical expertise and domain knowledge.

Specifically, by connecting an encoder-decoder network that accepts a task identifier to the parameters of another network responsible for processing inputs and generating predictions for that task, we enable a simultaneous search across the space of parametric functions and their associated parameter values.
Importantly, the resulting hyper-network allows for complete backpropagation and does not rely on the computation of higher-order derivatives for training, unlike alternative approaches \citep[see, e.g.,][]{finnModelAgnosticMetaLearningFast2017,liMetaSGDLearningLearn2017}.
This allows, even with relatively limited computational resources, to design a parametric model that is finely tuned for a specific family of tasks, using the allotted degrees of freedom per task to capture the variability between tasks.

Abstracting from the time-series nature of the data, the method belongs to a broader category of meta-learning and/or multi-task learning methods, depending on how exactly it is deployed in practice.
Meta-learning aims at designing/training a model based on multiple observed tasks so that it performs well when adapted with training data of yet unseen tasks from the same family, and then evaluated on the test data of that task.
In contrast, multi-task learning aims to achieve optimal performance on new data from tasks that were used for the initial training.
For an excellent review of these two closely related fields, please refer to \citet{hospedalesMetalearningNeuralNetworks2021}, \citet{huismanSurveyDeepMetalearning2021}, or \citet{zhangSurveyMultiTaskLearning2022}, respectively.
In this section, we will formulate the problem in terms of the meta-learning objective, as it is typically a more relevant paradigm for time-series forecasting.

Following the notation of \citet{hospedalesMetalearningNeuralNetworks2021}, we denote a task as $ \mathcal{T} = \{ \mathcal{D}_{train}, \mathcal{D}_{val} \}$.\footnote{
    \citet{hospedalesMetalearningNeuralNetworks2021} allow for a slightly more general setup in which the loss function may also differ across tasks. 
    However, this level of generality is not necessary for our purposes, so we suppress it for the ease of exposition.
}
This task consists of data generated by some DGP split into a training set $\mathcal{D}_{train} = \{ (x_{t}, y_{t})\}_{t=1}^{K}$ used for estimating model parameters, and a validation set $\mathcal{D}_{val} = \{ (x_{t}, y_{t})\}_{t=K+1}^{N}$ for which we aim to make predictions. 
The vector $x_{t}\in \mathbb{R}^{d_{x}}$ typically contains lagged values of $y_{t} \in \mathbb{R}^{d_{y}}$ or some transformation of these values.
Tasks are distributed according to an unknown distribution $p(\mathcal{T})$.

In the framework, a model consists of two components: the prediction function
\begin{equation}
    \hat{y}_{t}=f_{\omega}(x_{t};\hat{\theta})
\end{equation}
which outputs predictions of $y_{t}$ based on the predictors $x_{t}$, and the estimation function
\begin{equation}
    \hat{\theta} = \kappa_{\omega} (\mathcal{D}_{train})
\end{equation}
which outputs the vector of task-specific parameters $\hat{\theta}\in \Theta$ given the observations $\mathcal{D}_{train}$.
Both functions, $f_{\omega}(\cdot)$ and $\kappa_{\omega}(\cdot)$, are further parameterized by a vector of meta parameters $\omega \in \Omega$,  which are not directly dependent on the task $\mathcal{T}$ and generally encompass any prior decisions regarding the model (e.g., the choice of an appropriate model and its particular specification, estimation procedures, regularization techniques applied when estimating $\hat{\theta}$ etc.).
To clearly differentiate between the meta parameters $\omega$ and the task-specific parameters $\theta$, we will refer to the latter as \emph{mesa} parameters, following \citet{hubingerRisksLearnedOptimization2021}.

The quality of the model is assessed by the loss incurred on the evaluation set, denoted by $\mathcal{L}(\mathcal{D}_{val}; \hat{\theta}, \omega)$, with
\begin{equation}
    \mathcal{L}(\mathcal{D}; \hat{\theta}, \omega)  = \dfrac{1}{|\mathcal{D}|} \sum_{(x_{t}, y_{t})\in \mathcal{D}} \gamma\left( y_{t}, f_{\omega}(x_{t};\hat{\theta})\right)
\end{equation}
where the function $\gamma$ measures the discrepancy between $y_{t}$ and the prediction $\hat{y}_{t}$.
Typically, to align the process of finding the optimal parameters $\theta$, the estimation $\hat{\theta} = \kappa_{\omega} (\mathcal{D}_{train})$ is likewise performed by numerically minimizing the incurred loss over the training set:
\begin{equation}
    \hat{\theta} = \kappa_{\omega} (\mathcal{D}_{train}) \approx \underset{\theta \in \Theta}{\arg\min} \, \mathcal{L}(\mathcal{D}_{train};\theta, \omega).
\end{equation}

Oftentimes, the information contained in $\omega$ regarding which forecasting function $f_{\omega}(\cdot)$ to use and the most appropriate estimation function $\kappa_{\omega}(\cdot)$ is determined through expert judgment, based on informal prior knowledge regarding the task and/or ad-hoc hyperparameter tuning.
By considering a family of tasks distributed according to $p(\mathcal{T})$, we can formalize the problem of finding the most suitable model; $\omega$ such that, when observing $\mathcal{D}_{train}$ and adapting accordingly through $\hat{\theta}$, the expected performance on yet unobserved $\mathcal{D}_{val}$ will be minimized.
Formally:
\begin{equation}\label{eq:population_bilevel_opt}
    \begin{split}
        \omega^{*}  &=  \underset{\omega\in \Omega}{\arg\min} \, \underset{\mathcal{T}\sim p(\mathcal{T})}{\mathbb{E}} [  \mathcal{L}(\mathcal{D}_{val}; \hat{\theta}, \omega) ] \\
        \textnormal{s.t.:} \,\hat{\theta} &= \kappa_{\omega} (\mathcal{D}_{train}) \approx \underset{\theta \in \Theta}{\arg\min} \, \mathcal{L}(\mathcal{D}_{train};\theta, \omega).
    \end{split}
\end{equation}
Solving this problem is not feasible as the distribution $p(\mathcal{T})$ is unknown. 
However, given a collection of $M$ observed tasks $\lbrace\mathcal{T}^{(m)}\rbrace_{m=1}^{M}$, it is, at least in theory, possible to solve the finite sample equivalent of the problem instead:
\begin{equation}\label{eq:sample_bilevel_opt}
    \begin{split}
        \hat{\omega}  &=  \underset{\omega \in \Omega}{\arg\min} \,  \dfrac{1}{M}\sum_{m=1}^{M} \mathcal{L}(\mathcal{D}_{val}^{(m)}; \hat{\theta}^{(m)}, \omega)  \\
        \textnormal{s.t.:} \,\hat{\theta}^{(m)} &= \kappa_{\omega} (\mathcal{D}_{train}^{(m)}) \approx \underset{\theta \in \Theta}{\arg\min} \, \mathcal{L}(\mathcal{D}_{train}^{(m)};\theta, \omega).
    \end{split}
\end{equation}

\subsubsection{Architecture}

The bi-level optimization problem presented in Eq. \ref{eq:sample_bilevel_opt} is generally computationally demanding. 
It may be feasible to estimate $\{\hat{\theta}^{(m)}\}_{m=1}^{M}$ for a limited set of different model specifications $\Omega = \{\omega_{i}\}_{i=1}^{d_{\Omega}}$, and choose the model $f_{\omega_{i}}(\cdot)$ that yields the best out-of-sample performance over $\{\mathcal{D}^{(m)}_{val}\}_{m=1}^{M}$.
However, this approach quickly becomes untenable when the set $\Omega$ is large or even uncountable, for example, when considering a continuum of possible models rather than a limited set of predefined model specifications.

In addressing this problem, we adopt the following two simplifying assumptions:
\begin{itemize}[itemindent=2em]
    \item[\textbf{A1}:] The estimation function $\kappa_{\omega}(\cdot)$ outputs the global minimizer of the in-sample loss:
        \begin{equation}
            \forall \omega \in \Omega \,\forall  \mathcal{D}_{train}^{(m)} \in \left(\mathbb{R}^{d_{x}} \times \mathbb{R}^{d_{y}}\right)^{K} \exists! \theta^{*}\in \Theta:
        \end{equation}
        \begin{equation}
            \kappa_{\omega} (\mathcal{D}_{train}^{(m)})=\theta^{*} = \underset{\theta}{\arg\min} \, \mathcal{L}(\mathcal{D}_{train}^{(m)};\theta, \omega).
        \end{equation}
    \item[\textbf{A2}:] The training is conducted using a train-train split:
        \begin{equation}
            \begin{split}
                \hat{\omega} &=  \underset{\omega}{\arg\min} \, \dfrac{1}{M} \sum_{m=1}^{M}  \mathcal{L}(\mathcal{D}_{train}^{(m)}; \hat{\theta}^{(m)}, \omega)  \\
                \textnormal{s.t.:} \,\hat{\theta}^{(m)} &= \kappa_{\omega} (\mathcal{D}_{train}^{(m)}).
            \end{split}
        \end{equation}
\end{itemize}

Assumption A1 is pragmatically motivated by our aim, which is finding optimal parametric models.
This is in stark contrast to the widely popular family of meta-learning approaches derived from MAML \citep{finnModelAgnosticMetaLearningFast2017} that primarily concentrate on estimation routines.
There, $\omega$ typically represents the initial value of $\theta$ used in the estimation routine $\kappa_{\omega}$ or some additional information on how to adapt from $\theta$ (see, for example, \citet{finnModelAgnosticMetaLearningFast2017}, \citet{liMetaSGDLearningLearn2017}, and \citet{parkMetacurvature2019}).

Assumption A2 implies that the training is not conducted with the train-val split (i.e., with $\mathcal{D}_{val}^{(m)}$ in the outer optimization problem and $\mathcal{D}_{train}^{(m)}$ in the inner optimization problem), which is typical for meta-learning. 
Instead, it is done with a train-train split (i.e., using $\mathcal{D}_{train}^{(m)}$ in both the outer and inner optimization problems), as is common in multi-task learning.
In this setup, the validation datasets $\mathcal{D}_{val}^{(m)}$ are still utilized, but typically for early stopping of the training process rather than being directly included in the objective function.
This assumption is substantial because the training process, in this case, may not strictly correspond to way the model will be deployed in practice. 
That is, to the situation when observing a completely new task, $\mathcal{T}^{(M+1)}$, and being asked to adapt the model through $\theta^{(M+1)}$ based on $\mathcal{D}^{(M+1)}_{train}$ to predict $y$ in $\mathcal{D}^{(M+1)}_{val}$ while keeping $\omega$ fixed.
Despite this, it appears justifiable in light of studies that demonstrate that for meta-learning, the commonly adopted train-val split might not always be preferable to a simpler train-train split \citep{baiHowImportantTrainValidation2021} and that meta-learning and multi-task learning problems are closely connected \citep{wangBridgingMultitaskLearning2021}

The introduction of these assumptions substantially simplifies the optimization problem, as shown in the following proposition.

\begin{proposition} \label{prop:PropositionMain}
    Under assumptions A1 and A2, there exist functions $f(\cdot; \beta): \mathbb{R}^{d_x} \rightarrow \mathbb{R}^{d_y}$ parameterized by $\beta \in B$ and $g(\cdot; \omega):\Theta \rightarrow B$ parameterized by $\omega \in \Omega$, such that the solution of
    \begin{equation}\label{eq:sample_singlelevel_opt_alt}
        \begin{split}
            &\left\lbrace \hat{\omega} , \left\lbrace \hat{\theta}^{(m)} \right\rbrace_{m=1}^{M} \right\rbrace= \\
            &  \underset{\substack{\omega \in \Omega \\ \left\lbrace \theta^{(m)} \right\rbrace_{m=1}^{M}\in \Theta^{M}}}{\arg\min} \, \dfrac{1}{M} \sum_{m=1}^{M}  \dfrac{1}{K}\sum_{t=1}^{K} \gamma( y_{t}^{(m)}, f(x_{t}^{(m)};g(\theta^{(m)};\omega)))
        \end{split}
    \end{equation}
    coincides with the solution of the bilevel optimization problem introduced in Eq. \ref{eq:sample_bilevel_opt}.
\end{proposition}
The proposition demonstrates that under A1 and A2, the bilevel optimization problem in Eq. \ref{eq:sample_bilevel_opt} collapses to a much simpler, single-level optimization problem.
In this equivalent formulation, the model $f_{\omega}(\cdot, \theta^{(m)})$ is conveniently separated into two components: the \emph{base} model $f(\cdot; \beta^{(m)})$, parameterized by $\beta^{(m)}$, which processes features to generate predictions, and a \emph{meta} module $g(\theta^{(m)};\omega)$, which, based on the mesa parameter vector $\theta^{(m)}$, outputs the corresponding $\beta^{(m)}$. 
Thus, in effect, Proposition \ref{prop:PropositionMain} allows for a simultaneous search over both parametric functions $f_{\omega}$ and their corresponding mesa parameters $\{ \theta^{(m)} \}_{m=1}^{M}$.

To allow for maximal flexibility, we express both the base model $f(\cdot;\beta)$ and the meta module $g(\cdot;\omega)$ as feedforward neural networks. 
The total size of the network $f(\cdot;\beta)$, represented by $d_{\beta} = \textnormal{card}(\beta)$ , controls the level of complexity with which the predicted values $\hat{y}_{t}$ depend on the input $x_{t}$.
The size of the mesa parameters $d_{\theta}=\textnormal{card}(\theta)$ corresponds to the number of degrees of freedom allotted to each task $m$ and thus regulates the degree of globality/locality of the model.\footnote{
    If setting $d_{\theta}=1$ would still yield too much flexibility, it is also possible to further regularize the mesa parameters. 
    Allowing the regularization penalty to tend towards infinity renders the adaptation via $\theta$ ineffective, causing the model to collapse into a pure global model.
}
Finally, the size of the network $g(\cdot;\omega)$, represented by $d_{\omega} = \textnormal{card}(\omega)$, controls the nonlinearity of the model's response to mesa parameters $\theta^{(m)}$. 
Network $g$ does not necessarily have to be fully connected.
To reduce computational complexity, it is possible to leave some output nodes as orphaned constants, allowing the mesa parameters $\theta^{(m)}$ to affect only a part of the base model $f$, such as only its last layers.\footnote{
    This is motivated by the fact that adaptation predominantly occurs by altering the head of the network \citep{raghuRapidLearningFeature2019, linLearnEffectiveFeatures2020}.
}

\begin{figure*}[!htbp]
    \centering
    \includegraphics[width=0.8\linewidth]{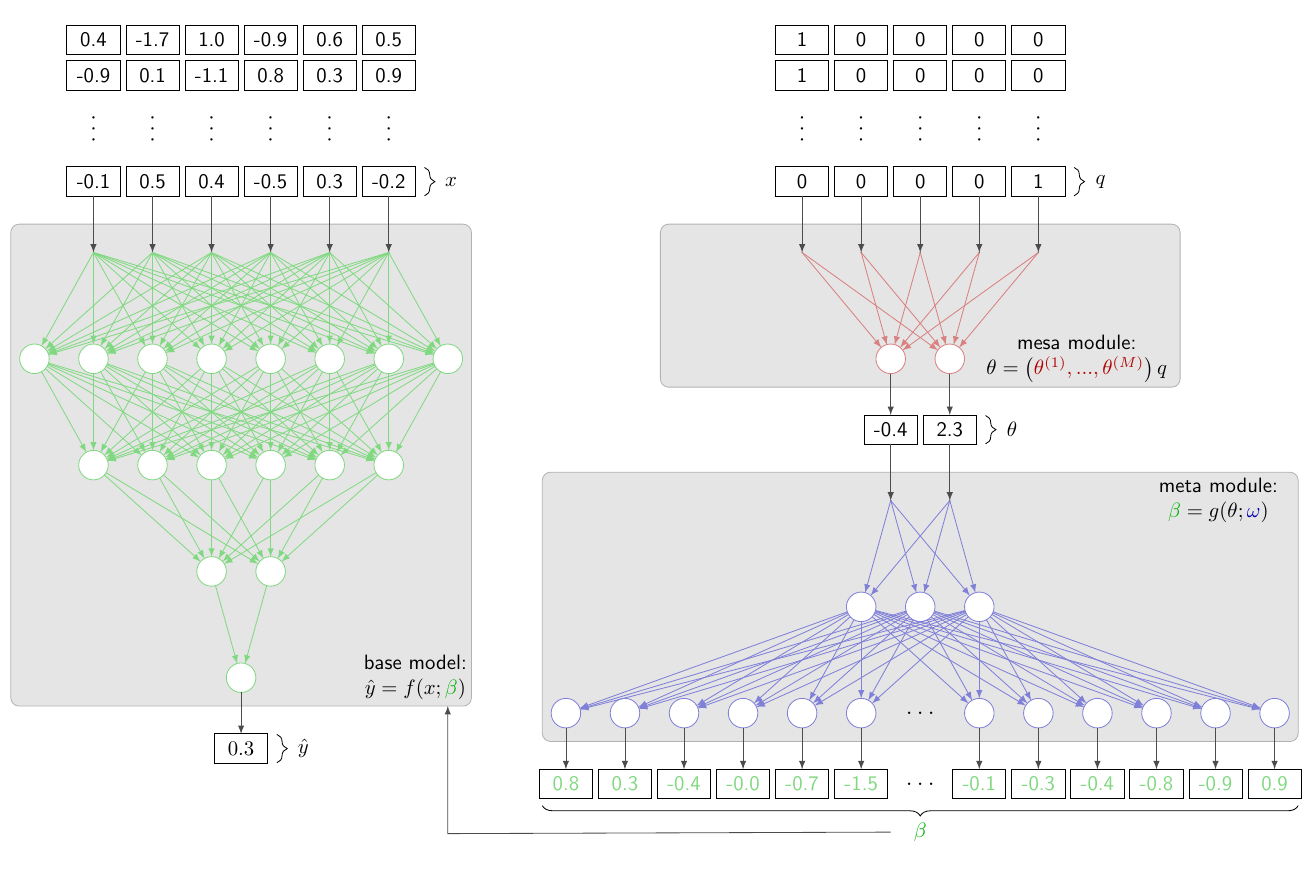}
    \caption{
        A diagram of the MtMs model for an illustrative example with 6 features and 5 tasks.
        The process of generating forecasts proceeds from the right to left. 
        First, a one-hot encoded vector $q$, denoting to which task the observation belongs, is multiplied by a matrix of mesa parameters $(\theta^{(1)},\,...\,,\theta^{(M)})$ to extract the corresponding task-specific mesa parameter vector $\theta$.
        This vector is then passed to the meta module $g(\theta; \omega)$ to generate task-specific parameters $\beta$ of the base model $f(x;\beta)$.
        Lastly, the network $f(x;\beta)$ is used to process the corresponding feature vector $x$ and generate the prediction $\hat{y}$.
    }
    \label{fig:Diagram}
\end{figure*}

Importantly, given that the optimization problem in Eq. \ref{eq:sample_singlelevel_opt_alt} is unconstrained and that both the meta parameters $\omega$ and the task-specific mesa parameters $\{\theta^{(m)}\}_{m=1}^{M}$ are optimized at the same level, the standard backpropagation techniques can be applied, considerably facilitating the training of the model.
When implementing the model, it is convenient to equivalently express the array of mesa parameters $\{\theta^{(m)}\}_{m=1}^{M}$ as a single neural network layer without any constants or nonlinearity.
This layer takes, as input, the one-hot encoding of the task $q=e_{m}$ and outputs the corresponding vector of mesa parameters  $\theta^{(m)} = [\theta^{(1)},\,...\,,\theta^{(M)}]q$.
The entire model can then be expressed as depicted in Figure \ref{fig:Diagram}.
For brevity, we will refer to the model simply as MtMs henceforth, emphasizing the simultaneous training of both global \textbf{m}e\textbf{t}a parameters $\omega$ and task-specific \textbf{m}e\textbf{s}a parameters $\{\theta^{(m)}\}_{m=1}^{M}$.

Despite being trained under the multi-task learning paradigm, the model can be deployed for both multi-task and meta-learning problems.
For multi-task learning, the model can be used as is without any further optimization.
By providing more data from an already observed task m $m$, predictions can be made using $f_{\hat{\omega}}(\cdot;\hat{\theta}^{(m)})=f(\cdot;g(\hat{\theta}^{(m)};\hat{\omega}))$ with the corresponding estimated mesa parameter vector $\hat{\theta}^{(m)}$.

For meta-learning applications, we leverage Proposition \ref{prop:PropositionMain}, which states that the solution $\hat{\omega}$ from Eq. \ref{eq:sample_singlelevel_opt_alt} can be, under simplifying conditions A1 and A2, interpreted as a parametric model $f_{\hat{\omega}}(\cdot;\theta)=f(\cdot;g(\theta;\hat{\omega}))$ that, out of all competing parametric models $\omega' \in \Omega$, delivers the smallest expected loss on a new task $\mathcal{T}^{(M+1)}$.
To predict on this previously unobserved task, it is therefore sufficient to perform optimization over the space of task-specific mesa parameters $\theta \in \Theta$:
\begin{equation}
    \hat{\theta}^{(M+1)} =  \underset{\theta \in \Theta}{\arg\min} \,\dfrac{1}{K}\sum_{t=1}^{K} \gamma( y_{t}^{(M+1)}, f(x_{t}^{(M+1)};g(\theta;\hat{\omega}))),
\end{equation}
while holding the model representation $\hat{\omega}$ fixed.

Note that this optimization is performed only in the low-dimensional space $\mathbb{R}^{d_{\theta}}$ and can be done using either backpropagation or conventional numerical optimization methods.
In this sense, it is completely analogous to finding parameters of any other parametric model.
The only difference is that the functional form of the model $f_{\hat{\omega}}(\cdot;\theta)$, as represented by $\hat{\omega}$, is not presupposed by the researcher but instead derived in a data-driven way specifically for the given family of prediction problems $p(\mathcal{T})$ in the initial meta-learning phase.
Similar to a conventional parametric model manually crafted by a human expert, the parameter vector  $\theta$ typically influence the prediction function $f_{\hat{\omega}}(\cdot;\theta)=f(\cdot;g(\theta;\hat{\omega}))$ in an interpretable way, as demonstrated in the applications later presented in this article (see \ref{section:sinusoidal}, \ref{section:m4}).
Furthermore, though not pursued in this article, the fact that $\hat{\theta}^{(M+1)}$ is an extremum estimator allows for inference regarding model parameters, provided that regularity conditions are met.

This method belongs to the strain of meta-learning research where hypernetworks/embeddings are used to perform adaptation to individual tasks at a lower-dimensional manifold of the parameter space \citep[see, e.g.,][]{leeGradientBasedMetaLearningLearned2018,zintgrafFastContextAdaptation2019,zhaoMetaLearningHypernetworks2020, flennerhagMetaLearningWarpedGradient2020, vonoswaldContinualLearningHypernetworks2022, navaMetaLearningClassifierFree2023, ramanarayananGeneralizingSupervisedDeep2023}.
The main point of differentiation is that in these studies, hypernetworks are generally used to facilitate fine-tuning of network weights while retaining the MAML paradigm of bilevel optimization, where the inner optimization is restricted to a few gradient steps due to computational constraints. 
In contrast to this approach of fine-tuning network weights, MtMs sidestep the bilevel problem formulation by virtue of assumption A2, which, in turn, allows one to interpret mesa parameters $\{\theta^{(m)}\}_{m=1}^{M}$ as global optimizers of some underlying parametric model crafted specifically for the family of tasks $p(\mathcal{T})$.
This is essential, as multistep task adaptation has been shown to be crucial in meta-learning \citep{linLearnEffectiveFeatures2020}. 
In this respect, the model is closely related to the seminal work of \citet{shamsianPersonalizedFederatedLearning2021}, where a similar architecture with a custom training algorithm (pFedHN) is proposed for the task of personalized federated learning.
To demonstrate that the applicability of MtMs is not limited only to financial time-series forecasting as in the case of M6, we also test its performance in two other environments, once under the meta-learning evaluation and once under the multi-task learning evaluation.
These additional demonstrations, available in the Appendices, allow us to compare the model's performance with established benchmarks.

In \ref{section:sinusoidal}, we apply MtMs to the problem of sinusoidal regression, a synthetic problem originally proposed by \citet{finnModelAgnosticMetaLearningFast2017} to test the performance of MAML.
This environment has since been frequently used to compare competing meta-learning methods.
In this environment, MtMs almost perfectly recover the underlying unobserved parametric model, and in out-of-sample evaluation, it substantially outperforms conventional meta-learning approaches such as MAML \citep{finnModelAgnosticMetaLearningFast2017}, Meta-SGD \citep{liMetaSGDLearningLearn2017}, MC \citep{parkMetacurvature2019}, and MH \citep{zhaoMetaLearningHypernetworks2020}.
This unparalleled performance stems from the fact that for the derived parametric model, represented by $\hat{\omega}$, prediction functions $f_{\hat{\omega}}(\cdot;\theta)=f(\cdot;g(\theta;\hat{\omega}))$ with varying mesa parameter $\theta$ almost perfectly mimic the various functions used to generate the data.
Consequently, when this derived parametric model is applied to a previously unseen task, only a handful of observations $\mathcal{D}_{train}^{(M+1)}$ are needed to unambiguously estimate $\hat{\theta}^{(M+1)}$ and hence precisely pinpoint the particular function used to simulate the given task.

In \ref{section:m4}, we apply MtMs under the multi-task learning paradigm to the time-series from the M4 forecasting competition, following the evaluation framework of \citet{montero-mansoPrinciplesAlgorithmsForecasting2021}.
A simple linear model localized via MtMs outperforms both the corresponding global model applied on pooled data and models localized via clustering for the majority of series.
Moreover, this very simple parametric linear model derived in a data-driven way through MtMs outperforms conventional widely used local models such as \texttt{ETS} \citep{hyndmanStateSpaceFramework2002a} and \texttt{auto.arima} \citep{hyndmanAutomaticTimeSeries2008a} on the majority of series.
Interestingly, the heterogeneity identified and modeled via $\theta$ seems to be primarily in the degree to which the time-series are persistent and exhibit seasonal patterns.

In the next subsection, we describe the primary application of MtMs discussed in this article: the forecasting challenge of the M6 competition.

\subsection[Application to M6 competition]{Application to the M6 competition\footnote{
        The model specification evolved slightly during the competition. 
        This section details the model's state as of the 12th and final submission. 
        For the evolution of the model, please refer to the accompanying repository.
    }
}\label{subsectition:mtms_application}

In the context of the forecasting challenge in the M6 competition, each task $m$ represents a single asset. 
The variable $y_{t}^{(m)}\in\mathbb{R}^{5}$ serves as an indicator for the quintile to which the returns of asset $m$ belong within the 4-week interval $t$, and $x_{t}^{(m)}$ is a feature vector used for prediction.

\subsubsection{Data augmention}

To enhance training stability and performance, we augment the dataset with assets beyond the 100 specified in the M6 universe.
Data augmentation is particularly advantageous for the MtMs model, as even if additional assets have substantially different DGPs from those in the M6 universe, these variations are likely to be absorbed by $\theta^{(m)}$.

We augment the original 50 stocks and 50 ETFs with an additional 450 stocks and 450 ETFs.
These assets are selected from a pool of assets with sufficient trading activity\footnote{Measured by the product of the daily traded volume and the closing price.} (must be at least 0.5 times the minimal trading activity observed in the M6 universe) and price history (must span from at least 2015 to the current date) to match the volatility observed in the M6 universe (the top 450 stocks/ETFs with the highest likelihood of their volatility being observed among the stocks/ETFs in the M6 universe are selected).
Finally, the additional 450 stocks and 450 ETFs were randomly divided into 9 additional M6-like universes in order to compute quintiles $y_{t}^{(m)}$ of returns.\footnote{
    Note that computing quintiles based on all 900 additional assets at once does not generally align with the original objective.
}

In addition to augmentation across the dimension $M$, we calculate quintiles $y_{t}^{(m)}$ and features $x_{t}^{(m)}$ for 4-week intervals shifted by 1, 2, and 3 weeks relative to the actual start of the competition (2022-03-07).
Assuming the time-series $y_{t}^{(m)}$ and $x_{t}^{(m)}$ are stationary, such augmentation does not alter the objective in any way and allows us to effortlessly quadruple the amount of data per asset $m$, further enhancing the stability of the training process.

\subsubsection{Features}

As features $x_{t}^{(m)}$, we utilize an indicator for whether a given asset is an ETF, its own lagged 4-week returns and volatilities (up to lag 7), and an array of technical trading indicators from the TTR package \citep{ulrichTTRTechnicalTrading2021}, calculated based on historical prices.
We opt for TTR because it offers a unified interface, allowing us to generate a diverse set of features programmatically without requiring manual adjustments. 
A complete list of all 81 features is provided in Table \ref{tab:features}.\footnote{
    Some indicators are multivariate and/or are computed with different lengths of the rolling window.
    The feature selection process involved initially training an XGBoost model \citep{chenXgboostExtremeGradient2023} using all available technical trading indicators from TTR and subsequently pruning the least important features.
}
Finally, we impute missing values with medians, and normalize the features to zero mean and unit variance.

\subsubsection{Model \& training}

\begin{figure*}[!htbp]
    \centering
    \includegraphics[width=0.8\linewidth]{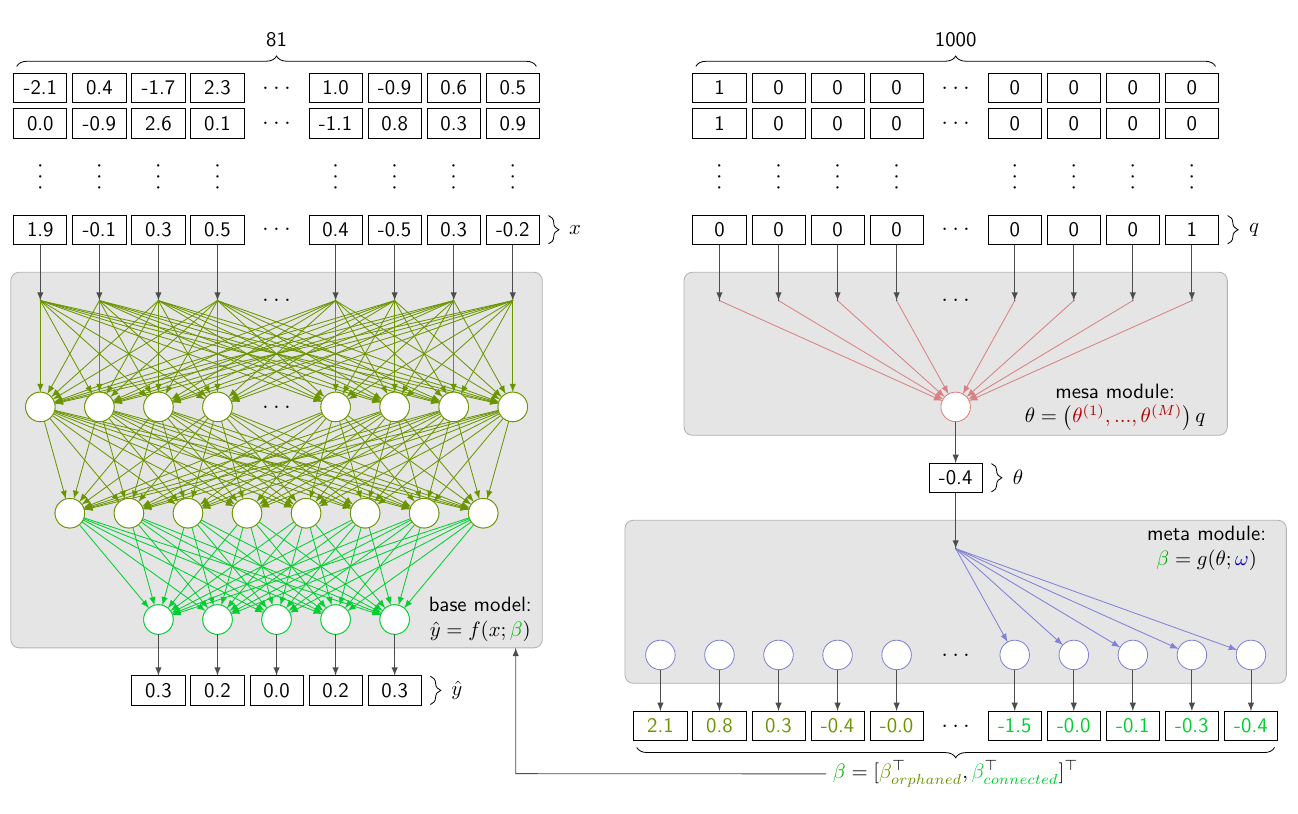}
    \caption{
        A diagram of the MtMs model applied to M6.
        In the case of M6, there are 1000 tasks/assets (100 specified by the organizers and 900 from the additional 9 auxiliary M6-like datasets). 
        Each asset is allotted one univariate mesa parameter $\theta$, which, through the meta module $g(\theta; \omega)$, determines the parameters $\beta$ of the network $f(x;\beta)$. 
        This network then processes the corresponding feature vector $x$ to generate the prediction $\hat{y}$.
        The meta module $g(\theta; \omega)$ is a trivial single-layer neural network that connects $\theta$ to the weights and biases of the last layer of the network $f$; $\beta_{connected}$.
        The remaining nodes corresponding to parameters $\beta_{orphaned}$ are not influenced by $\theta$ and are hence constant across all tasks/assets.
    }
    \label{fig:Diagram_specific}
\end{figure*}

The base model $f(\cdot;\beta)$ is a feedforward neural network comprising two hidden layers with 32 and 8 units, featuring leaky ReLU nonlinearity and a dropout rate of 0.2. 
The output layer has 5 units and utilizes a softmax transform.
The meta module $g(\cdot;\omega)$ is a trivial feedforward network with no hidden layers or nonlinearity.
One mesa parameter ($d_{\theta}=1$) is allotted to each asset, influencing the weights and biases of the final layer in $f(\cdot; \beta)$.
The architecture of the entire model is displayed in Figure \ref{fig:Diagram_specific}.

To train the model, we utilize data from 2000 to 2022 for training, reserving the remaining data for testing. 
Given the high sensitivity of hypernetworks to their initialization \citep{beckHypernetworksMetaReinforcementLearning2023}, our training process consists of two steps.
In the first step, the base model is trained on pooled data without taking into account which data belongs to which task. 
This training is conducted under the RPS loss using the Adam optimizer with a learning rate of 0.01, a minibatch size of 200, and early stopping.

In the second step, the trained weights from the first step serve as an initialization for the bias of the meta module $g(\cdot; \omega)$. 
Meanwhile, the weights of the meta module are initialized uniformly on the interval $\left[-1, 1\right]$, and mesa parameters $\{\theta^{(m)}\}_{m=1}^{M}$ are set to 0.
This means that the optimization begins from a point where the MtMs model is already proficient at predicting $y_{t}^{(m)}$, and the objective now is primarily to capture any systematic differences among the DGPs of individual assets through the mesa parameters $\{\theta^{(m)}\}_{m=1}^{M}$.
The optimization is carried out iteratively using the Adam optimizer, with gradually decreasing learning rates (values $\splitatcommas{\{0.01,0.001,0.001,0.0005,0.0003,0.0001,0.00005\}}$), minibatches consisting of 100 randomly selected assets and early stopping.
We employ this repeated training scheme because MtMs can be challenging to train, with the optimizer often struggling to adjust the model weights for improved test loss on the initial attempt. 
Multiple iterations are typically required.
Finally, to make predictions, we can readily employ estimated mesa parameters $\hat{\theta}^{(m)}$ corresponding to the original asset universe without any further training (i.e., multi-task learning deployment).

\subsubsection{Post-processing \& predictions}\label{subsection:mtms_predictions}

Although the rankings of individual assets are intrinsically related (with exactly 20 assets belonging to each quintile within each universe), we choose to disregard this dependence and submit predictions $\hat{y}_{t}^{(m)}=f(x_{t}^{(m)},g(\widehat{\theta}^{(m)};\widehat{\omega}))$ without any post-processing or further adjustments.\footnote{
    The only exceptions are the predictions for the DRE stock during submissions 10-12. 
    After DRE stock was acquired by PLD, it exhibited zero price changes from that point forward. 
    To address this, we overrode the predictions with observed frequencies with which a hypothetical asset with zero returns would belong to individual quintiles.
}
While harmonizing the predictions could potentially yield performance improvements, we did not pursue this as the universe's size of 100 assets is adequate to ensure that $y_{t}^{(m)}$ is at least approximately unrelated in this regard.

\begin{figure}[!htbp]
    \centering
    \includegraphics[width=.90\linewidth]{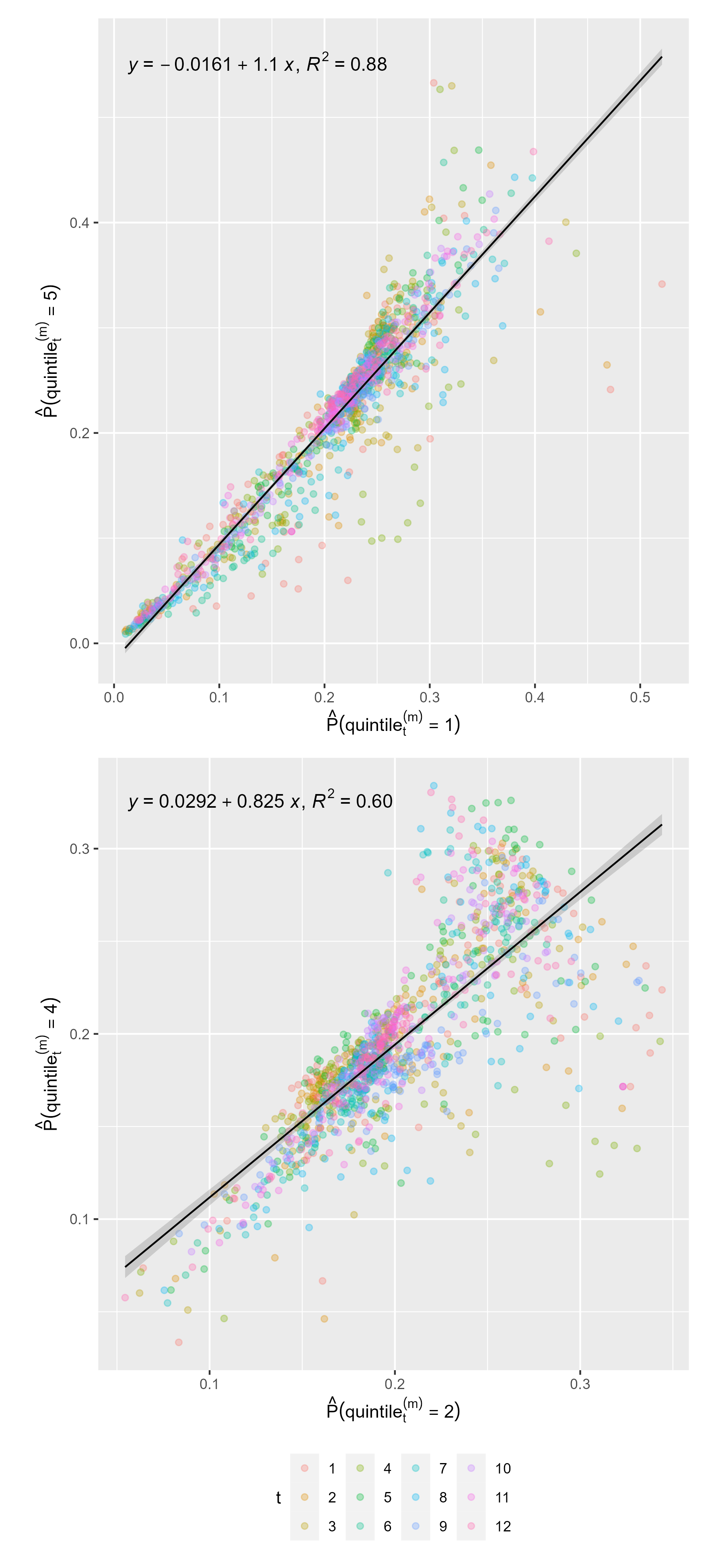}
    \caption{
        Predicted probabilities of the 1st quintile plotted against the probabilities of the 5th quintile (upper panel) and predicted probabilities of the 2nd quintile plotted against the probabilities of the 4th quintile (lower panel).
    }
    \label{fig:quintile_plots}
\end{figure}

Interestingly, despite performing well when measured by RPS loss ($0.15689$ over the duration of the competition), the predictions generated by the model contain surprisingly little directional information.
This severely limits their practical utility for forming investment portfolios, except for risk management purposes.
Figure \ref{fig:quintile_plots} displays the predicted probabilities of the 1st (resp. 2nd) quintile plotted against predicted probabilities of the 5th (resp 4th) quintile for individual assets throughout the competition.
Predictions generally traverse along the diagonal line, implying that an increased probability of exceptionally good performance, relative to other assets, is accompanied by an increased probability of exceptionally poor performance, and vice versa, thus failing to provide any clear recommendations on which positions to take.
This finding appears to align with the efficient market hypothesis \citep[see, e.g.,][]{malkielReflectionsEfficientMarket2005}, which posits that it is impossible to achieve abnormal returns based on information contained in the price history.

Disentangling the precise causes of the model's relatively good performance is challenging. 
However, the training metrics suggest that the considered assets are relatively homogenous. 
The most significant improvements over naive predictions were achieved through joint training, with adaptation playing a secondary role. 
MtMs nonetheless still provided the advantage of using a much broader universe of assets for training without concerns about their dissimilarity to the assets specified by the organizers.

\section{Investment challenge}\label{section:investment_challenge}

Before the start of the competition, we made several attempts to systematically transform quintile predictions into portfolio weights.
These attempts ranged from a parametric approach that combined information about the marginal distributions with the correlation structure using copulas to a fully nonparametric approach. 
However, perhaps unsurprisingly, given the notorious difficulty of achieving abnormal returns and given the results displayed in Figure \ref{fig:quintile_plots}, none of these approaches passed backtesting.

In general, the predictability of quintiles of rank does not necessarily imply predictability of expected returns; many DGPs for asset returns with identical means are compatible with non-uniform and predictable quintiles. 
Even the minor asymmetries in quintile predictions\footnote{I.e., a situations in which $\hat{P}(quintile_{t}^{(m)}=1) \neq \hat{P}(quintile_{t}^{(m)}=5)$ or $\hat{P}(quintile_{t}^{(m)}=2) \neq \hat{P}(quintile_{t}^{(m)}=4)$.} occasionally observed in Figure \ref{fig:quintile_plots} are not necessarily indicative of mean predictability.
They could, and likely are, given our inability to capitalize on them, caused by a varying degree of asymmetry in the distribution of returns across different assets.

Given the failure to produce investment decisions that consistently attained abnormal returns, we opted to use the investment challenge as ancillary to the prediction challenge.
The decisions were made on a discretionary basis and were primarily guided by two principles.
First, the portfolio weights were scaled to gain a small but certain advantage.
Second, the signs of the positions were strategically altered depending on the current ranking to improve the chances of securing a good enough rank in the duathlon challenge.

\subsection{Scaling}\label{subsection:scaling}

Given a collection portfolio weights $\{w_{t}^{(m)}\}_{m=1}^{M}$ for $M$ assets submitted at time $t$, we can consider the decomposition:
\begin{equation}
    w_{t}^{(m)}=\underbrace{\alpha_{t}}_{\sum_{m'=1}^{M}|w_{t}^{(m')}|}  * \underbrace{\tilde{w}_{t}^{(m)}}_{ \alpha_{t}^{-1}w_{t}^{(m)} }.
\end{equation}
Here, $\tilde{w}_{t}^{(m)}$ is a scaleless portfolio weight, and $\alpha_{t}$ is the overall scaling factor at time $t$.

The objective function \citep[see][]{makridakisM6FinancialDuathlon2022} involves a log transform, which, due to its concavity, penalizes extreme returns more severely. 
Therefore, irrespective of $\{\tilde{w}_{t}^{(m)}\}_{m=1}^{M}$, it is desirable to set $\alpha_{t}$ as small as possible to minimize the dispersion of returns prior to standardization.
Setting $\alpha_{t}=0.25$ results in a very modest but certain advantage over $\alpha_{t}=1$. 
In the case of an equal-weighted long portfolio, this amounts to approximately 0.07 IR per 4-week interval (see Fig. \ref{fig:scaling_effects}).

\subsection{Strategic positions}\label{subsection:strategic}

To maximize the probability of securing the top rank, it is desirable to take more risky positions when one ranks poorly in the public leaderboard, attempting to improve otherwise hopeless positions.
Conversely, more conservative positions might be warranted if one already holds a sufficiently good rank and only wishes to maintain it.
However, because the objective is defined in terms of risk-adjusted returns, it is challenging to directly control risk by forming portfolios with varying degrees of return variability.

To circumvent this issue, we attempted to leverage the competitive nature of the competition, where only relative performance matters, and the fact that participants were primarily taking long positions.\footnote{
    Despite the submitted positions of individual teams being private, the number of participants submitting predominantly long positions can be approximately inferred from the public leaderboard by observing for how many participants their monthly IR is of the same sign as that of the equal-weighted long benchmark portfolio in each month.
    A slightly more rigorous approach, though not pursued at the time of the competition, is to match properties of simulated IR with the actually observed IR in the leaderboard via the method of simulated moments \citep{mcfaddenMethodSimulatedMoments1989}, as in \citet{stanekNoteM6Forecasting2023a}.
}
Let us denote the number of long positions in the portfolio at time $t$ as $n^{+}_{t}$ and the number of short positions as $n^{-}_{t}$.
By varying the proportion of short positions in the portfolio $n^{-}_{t}/(n^{+}_{t}+n^{-}_{t})$, one can control the extent to which returns of the submitted portfolio would be negatively or positively correlated with the returns of other participants at large.
This allowed us to either increase the chances of rapidly climbing or descending the leaderboard (if achieving a top rank was otherwise unlikely) or reduce the risk of losing an already satisfactory ranking to a competitor (if a sufficiently good rank had already been achieved).

A formalization of this type of approach as a dynamic programming problem, along with an analysis of its performance, can be found in \citep{stanekNoteM6Forecasting2023a}.
The simulations suggest that employing such an adversarial portfolio strategy can significantly improve the likelihood of achieving a favorable rank.
This effect is particularly notable for the highest rankings; the probability of securing the 1st place is approximately 3 times higher than expected by chance, comparable to that of a participant consistently generating double the market returns.
The advantage for less extreme placements is less pronounced, with the probability of securing the 20th place or better being approximately 1.5 times higher than expected by chance.
However, this improvement comes at the expense of negative expected returns and a disproportionately higher probability of achieving an extremely poor rank due to its reliance on aggressive shorting and martingale-like risk-taking.

\subsection{Investment decisions}\label{subsection:decisions}

\begin{figure}[!htbp]
    \centering
    \includegraphics[width=.90\linewidth]{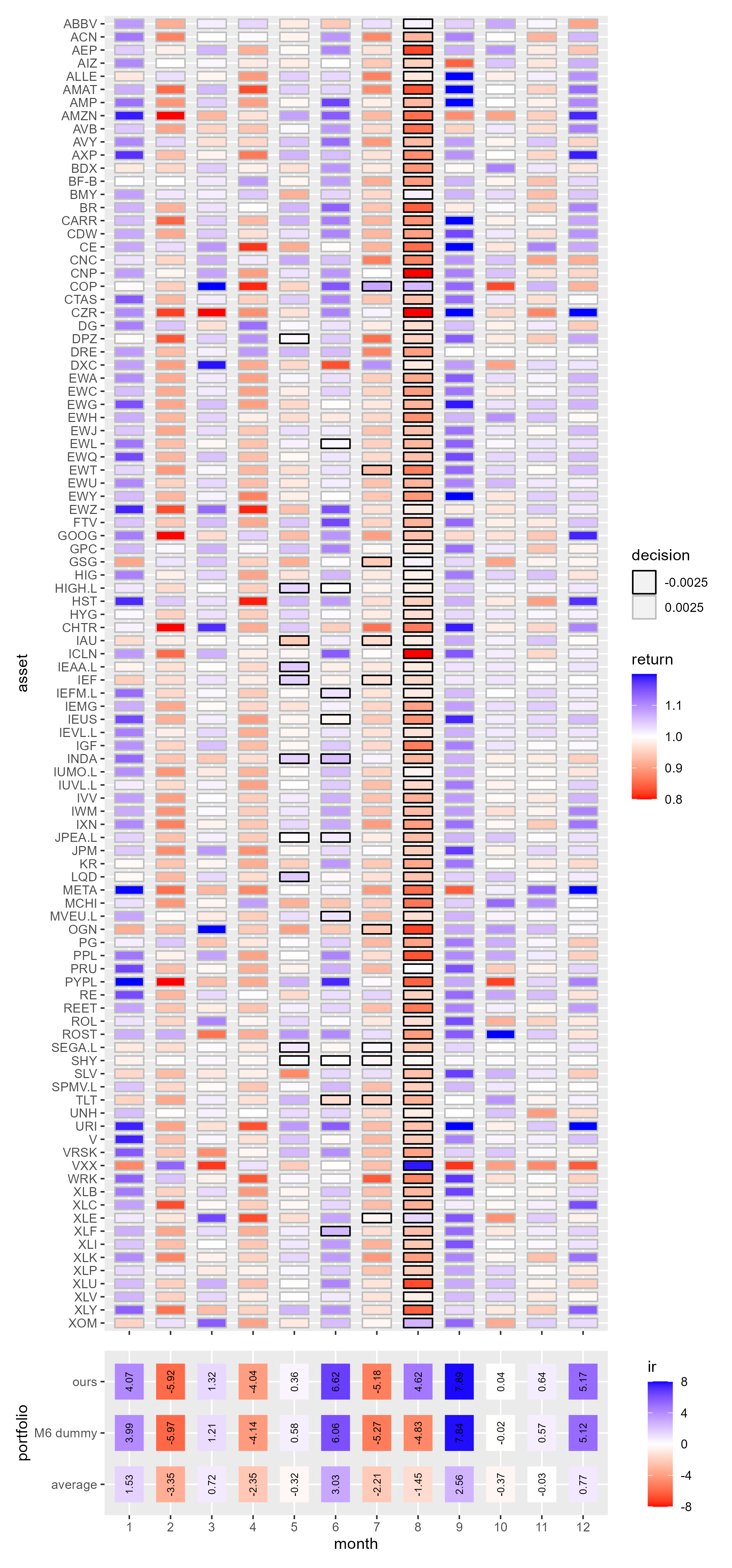}
    \caption{
        Portfolio weights and the overall performance of the portfolio across individual submissions. Performance of the \emph{M6 dummy} portfolio and the average performance of participants for comparison.
    }
    \label{fig:portfolio_weights}
\end{figure}

Throughout the competition, we maintained $|w_{t}^{(m)}|=0.0025$, with only the signs of the weights being manipulated. 
Figure \ref{fig:portfolio_weights} presents a comprehensive overview of the submitted portfolio weights, returns of individual assets, and the overall portfolio returns for our submissions, along with the returns of reference portfolios for comparison.

In submissions 1-4, $n^{-}_{t}$ was set to 0.
Starting from submission 5, it became apparent that scaling alone would be insufficient to secure a top rank, and that some risk indeed needed to be taken.
During submissions 5-7, $n^{-}_{t}$ was set to 10 to induce greater dispersion between our returns and those of competitors.
As the strategy described in section \ref{subsection:strategic} provides no guidelines regarding which particular assets one should short, we selected them using an ad hoc rule according to the difference of predicted probabilities of the lowest and the highest quintile. 
Unsurprisingly, these choices turned out no better than if one would be selecting at random (see Fig. \ref{fig:portfolio_weights}).
In response to the gradually worsening position in the duathlon leaderboard, $n^{-}_{t}$ was set to 100 in submission 8.
Coincidentally, returns during that period were predominantly negative, which substantially improved our ranking.
For the remainder of the competition, submissions 9-12, $n^{-}_{t}$ was set again to 0 to induce a positive correlation with returns of possible contenders and hence minimize the probability of losing an already sufficiently good rank.

While the strategy described in Subsection \ref{subsection:strategic} can, to some degree, increase the probability of securing a relatively good rank (and likewise, a bad rank), it should be emphasized that the realized rank is, in our case, still predominantly the result of mere chance.
In particular, by optimally selecting $n^{-}_{t}$, one may, in a stylized environment, increase the chances of ranking in the top 20 in the investment challenge (and, therefore, securing 1st place in the duathlon, with the forecasting challenge results held constant) from 0.12 to 0.19 (upper estimate, see \citep{stanekNoteM6Forecasting2023a}). 
While a non-negligible improvement, it is still very far from a guaranteed outcome.

\section{Conclusions}\label{section:conclusions}

We describe our methods for the forecasting and investment challenges in the M6 forecasting competition, which secured 4th and 6th place, respectively, ultimately winning 1st place in the duathlon.

For the forecasting challenge, we employed a meta-learning/multi-task learning model based on hypernetworks.
This approach enables the creation of a parametric model specifically optimized for a particular family of prediction problems.
In this way, the model's parameters are used to capture any between-task variability, while features of the DGP that are approximately invariant across tasks are learned from the pooled data. 
The resulting quintile predictions yield good performance in terms of RPS loss. 
However, they seem to provide no useful information about the expected values of returns.

For the investment challenge, we attempted to increase our chances of securing a good rank by employing a simple strategy, which aims to maximize/minimize the dispersion of the differences between our IR and the IRs of other competitors, depending on our current position in the public leaderboard.

Finally, despite the promising results, it's important to exercise caution and not place undue focus on top-performing approaches.
The performance of any approach, despite being measured over the span of one year, is still partially obscured by statistical noise, making it difficult to determine if one model truly outperforms another in expectation.
The top-performing participants, by the very fact that they secured good rankings, were likely more fortunate than others.
In the case of the investment challenge, our own simulations suggest that achieving the 6th rank would likely not be repeated if the competition were held again.
The same caveat also applies to our results in the forecasting challenge, although here, the fact that the MtMs model also outperforms state-of-the-art meta-learning approaches on the sinusoidal regression problem (\ref{section:sinusoidal}) and that it performs excellently on M4 (\ref{section:m4}) certainly shows great promise.

An evident and natural extension of this work is to evaluate MtMs on other widely recognized meta-learning problems, such as few-shot image classification (using datasets like Omniglot \citep{lake2011one} and Mini-ImageNet \citep{ravi2016optimization}) or reinforcement learning (2D navigation and locomotion \citep[see, e.g.,][]{finnModelAgnosticMetaLearningFast2017}).

\section{Acknowledgments}\label{section:acknowledgments}
Our sincere appreciation goes to the M6 organizers for their dedication in coordinating this challenging year-long competition.
We also extend our gratitude to Spyros Makridakis, Evangelos Spiliotis, and Fotios Petropoulos for their insightful comments, which have greatly improved this article.

\appendix

\section{Sinusoidal regression}\label{section:sinusoidal}
To evaluate the potential of the MtMs to find the most appropriate parametric model for a given family of prediction problems, we consider a simulation exercise originally proposed by \citet{finnModelAgnosticMetaLearningFast2017} to test the performance of MAML.
Since then, this environment has frequently been used to compare competing meta-learning methods.
The problem involves predicting $y \in \mathbb{R}$ based on $x \in \mathbb{R}$, where each task's data follows a sine wave with randomly sampled amplitude and phase.
Importantly, phase and amplitude are not observed directly, making it challenging to infer the underlying family of functions due to the limited number of observations per task and substantial heterogeneity across tasks.
In particular, the tasks $ \mathcal{T}^{(m)} = \{ \mathcal{D}_{train}^{(m)}, \mathcal{D}_{val}^{(m)} \}  $ are generated according to the following DGP:\footnote{
    As the sinusoidal regression is a cross-sectional exercise, we index individual observations by $i$ rather than $t$ to highlight that they are conditionally IID.
}
\begin{equation}\label{eq:sinDGP}
    \begin{split}
        A^{(m)} &\sim U(0.1,5)\\
        b^{(m)} &\sim U(0,\pi)\\
        x_{i}^{(m)}|A^{(m)}, b^{(m)} &\sim U(-5,5) \\
        y_{i}^{(m)}|x_{i}^{(m)}, A^{(m)}, b^{(m)} &= A^{(m)} * sin(x_{i}^{(m)}+b^{(m)})
    \end{split}
\end{equation}
The goal is to find the best model that can predict $y_{i}^{(m)}$ based on $x_{i}^{(m)}$ for $i > K$ after observing only $K$ observations $\mathcal{D}^{(m)}_{train}$, as measured by the mean squared error:

\small
\begin{equation}
    \begin{split}
        \mathcal{L}(\mathcal{D}_{val}^{(m)};\hat{\theta}^{(m)},\omega) &= \dfrac{1}{N-K}  \sum_{i = K+1}^{N} (y_{i}^{(m)}-f_{\omega}(x_{i}^{(m)};\hat{\theta}^{(m)}))^{2} \\
        \textnormal{s.t.:} \,\hat{\theta}^{(m)} &= \kappa_{\omega} (\mathcal{D}_{train}^{(m)})
    \end{split}
\end{equation}
\normalsize

For fair comparison, we follow \citet{finnModelAgnosticMetaLearningFast2017} and set the base model to be a feedforward neural network with two hidden layers of size $40$ and ReLU non-linearities.
The number of mesa parameters, $d_{\theta}$, is set to $2$ and the meta module $g(\cdot;\omega)$ is a simple fully connected feedforward network with no hidden layers or non-linearities.
MtMs is first trained via the Adam optimizer with a learning rate of 0.001 and minibatches of 100 tasks on $M=1000$ randomly generated tasks\footnote{Fewer than the 70,000 tasks originally used in \citet{finnModelAgnosticMetaLearningFast2017} and in the follow-up studies.} with $K$ observations.
After the initial meta-learning phase to identify the most suitable parametric model, the resulting model with fixed $\hat{\omega}$ is evaluated on 600 previously unseen tasks.
For each new task $m'$ with $K$ observed datapoints $\mathcal{D}_{train}^{(m')}$, a two-dimensional optimization for $\hat{\theta}^{(m')}\in \mathbb{R}^{2}$ is performed (using the Adadelta optimizer with a learning rate of 0.001) in order to find the mesa-parameter vector most suitable for the sampled task.
The estimated $\hat{\theta}^{(m')}$ is then used to make predictions on $\mathcal{D}_{val}^{(m')}$.
Other simulation details follow \citet{zhaoMetaLearningHypernetworks2020} and are available in the replication repository.

Table \ref{tab:sinLosses} shows the mean squared error achieved by the MtMs for $5$-shot learning and $10$-shot learning.
For comparison, we include the losses of commonly used meta-learning methods on this task (the performance of competing methods is taken from \citet{parkMetacurvature2019} and \citet{zhaoMetaLearningHypernetworks2020}).
The proposed MtMs model outperforms all benchmark methods by an order of magnitude for both $5$-shot learning and $10$-shot learning of the sinusoidal task.
In fact, the losses are in both cases very close to the theoretical minimum of $0$, indicating that the MtMs is capable of recovering the data-generating process to such a degree that, when faced with only as few as $5$ observations $(x_{i}^{(m')},y_{i}^{(m')})$ from task $m'$, it is able to almost perfectly infer $y_{i}^{(m')}$ as a function of $x_{i}^{(m')}$ for the whole range $[-5,5]$.

\begin{table}[!htbp]
    \fontsize{7}{7}\selectfont
    \centering 
    \begin{tabular}{l c c} 
        \toprule
        Method                                             & $K=5$               & $K=10$              \\
        \midrule
        MAML \citep{finnModelAgnosticMetaLearningFast2017} & $0.686^{\pm 0.070}$ & $0.435^{\pm 0.039}$ \\ 
        LayerLR \citep{parkMetacurvature2019}              & $0.528^{\pm 0.068}$ & $0.269^{\pm 0.027}$ \\
        Meta-SGD \citep{liMetaSGDLearningLearn2017}        & $0.482^{\pm 0.061}$ & $0.258^{\pm 0.026}$ \\
        MC1 \citep{parkMetacurvature2019}                  & $0.426^{\pm 0.054}$ & $0.239^{\pm 0.025}$ \\
        MC2 \citep{parkMetacurvature2019}                  & $0.405^{\pm 0.048}$ & $0.201^{\pm 0.020}$ \\
        MH \citep{zhaoMetaLearningHypernetworks2020}       & $0.501^{\pm 0.082}$ & $0.281^{\pm 0.072}$ \\
        MtMs (ours)                                        & $\bf 0.022^{\pm 0.003}$ & $\bf 0.014^{\pm 0.001}$ \\
        \bottomrule 
    \end{tabular}
    \caption{
        Losses for sinusoidal task\\
        Mean squared errors and corresponding $95\%$ confidence intervals for different meta-learning methods.
        Bold text indicates the best-performing model.
    } 
    \label{tab:sinLosses} 
\end{table}

\begin{figure}[!htbp]
    \centering
    \includegraphics[width=.99\linewidth]{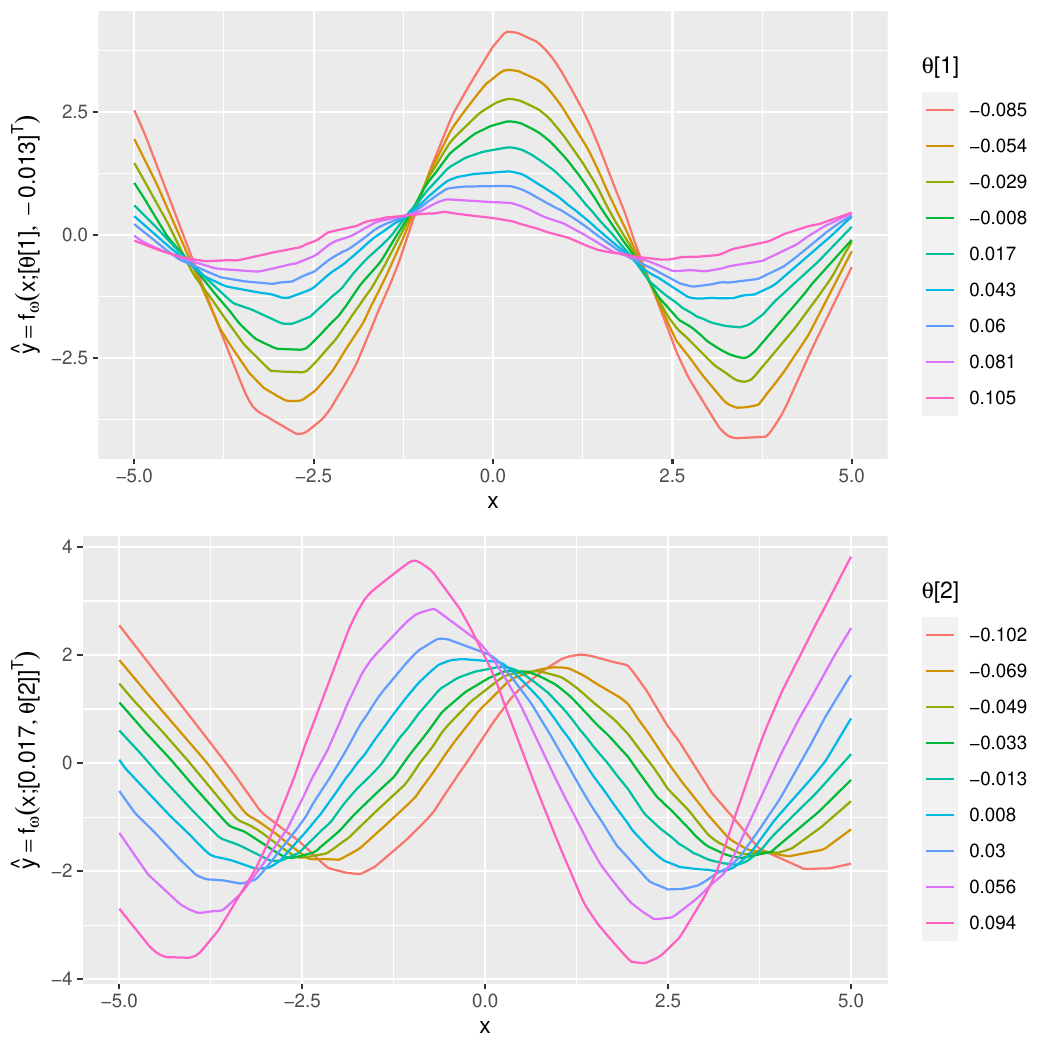}
    \caption{
        MtMs predictions for sinusoidal task ($K=5$)\\
        Plots of $f_{\omega}(x;\theta)$ as a function of $x$ for different values of the mesa parameter vector $\theta$.
        In the upper panel, the first mesa parameter $\theta[1] $ varies while $\theta[2] $ is fixed to its median value $-0.013$.
        In the lower panel, the second mesa parameter $\theta[2] $ varies while $\theta[1] $ is fixed to its median value $0.017$.
    }
    \label{fig:sinPredictionFunctions}
\end{figure}

The fact that the sinusoidal regression problem is univariate allows us to conveniently visualize the type of parametric model learned from the data during the initial meta-learning phase.
Figure \ref{fig:sinPredictionFunctions} shows predictions of the model $f_{\omega}(x;\theta)$ as a function of $x$ for different values of mesa-parameters $\theta$.
As is apparent from Figure \ref{fig:sinPredictionFunctions}, the plotted prediction functions closely resemble different sine waves, indicating that MtMs is indeed capable of correctly inferring that each generated task follows a sine function with varying phase and amplitude.
However, the mesa parameters $\theta=[\theta[1],\theta[2]]^{\top} $ explaining the variability between tasks do not directly correspond to the amplitude $A$ and phase $b$.
Instead, $\theta[1]$ regulates the amplitude (negatively), but to a lesser degree, it also regulates the phase (positively), while $\theta[2]$ primarily regulates the phase (positively) and, to a lesser degree, it also regulates the amplitude.
This is not surprising, as there are infinitely many parametric models that are observationally equivalent to the DGP described in Eq. \ref{eq:sinDGP}.
In particular, any two vectors in $\mathbb{R}^2$ that are linearly independent are capable of spanning the whole space of $[b,A]$ just as well as the basis vectors used in Eq. \ref{eq:sinDGP}.
The MtMs hence generally converges to one of these equivalent parametrizations, not necessarily to the exact same parametrization used to simulate the data.

Admittedly, the sinusoidal regression problem is relatively favorable to MtMs because the data are generated deterministically using a clearly defined low-dimensional model, and MtMs is, at its core, a method for recovering unknown parametric models.
Nonetheless, the fact that MtMs is capable of achieving unparalleled, near-oracle performance even when using fewer tasks than previous studies clearly demonstrates its capacity to identify and model the latent variability among the observed tasks.

\section{M4}\label{section:m4}

To assess the ability of the MtMs model to localize global time-series forecasting models in more general settings than those encountered in M6, we also perform an extensive evaluation on the data from the M4 forecasting competition \citep{makridakisM4Competition1002020}.
We follow the evaluation framework of \citet{montero-mansoPrinciplesAlgorithmsForecasting2021}, who demonstrated the surprising performance of simple global models, including a simple pooled OLS with lagged values of the time series as regressors.
We extend this forecasting exercise by exploring the extent to which the performance of the pooled OLS can be improved through localization via time-series clustering and MtMs.

Following \citet{montero-mansoPrinciplesAlgorithmsForecasting2021}, we focus on time-series with yearly, quarterly, monthly, and weekly frequencies (for forecast horizons of 6, 8, 18, and 13, respectively, using the recursive forecasting scheme) and use the MASE loss function with scaling applied as a preprocessing step.
The feature vectors contain lagged values of the given time-series: $x_{t}^{(m)}=[y_{t-1}^{(m)}, y_{t-2}^{(m)}, \ldots,y_{t-d_{x}}^{(m)}]^{\top}$.
For a time-series $m$ of length $d_{m}$, the design matrix is defined as $X^{(m)}=[x_{d_{x}+1}^{(m)},\,x_{d_{x}+2}^{(m)},\,\ldots,\,x_{d_{m}}^{(m)}]^{\top}$, and the dependent variable vector is $y^{(m)} = [y_{d_{x}+1}^{(m)},\,y_{d_{x}+2}^{(m)},\,\ldots,\,y_{d_{m}}^{(m)}]^{\top}$.
By stacking $\{y^{(m)}\}_{m=1}^{M}$ and $\{X^{(m)}\}_{m=1}^{M}$, one can obtain the design matrix and dependent variable vector for pooled regression to estimate the pooled $\beta$ for all time-series of a given frequency.
Likewise, after performing time-series clustering to account for heterogeneity across series, one can obtain $\beta$ for each cluster of similar series.

To mimic the same settings with MtMs, we set $f(x;\beta^{(m)})=x^{\top}\beta^{(m)}$, and define $g(\theta;\omega)$ as a simple neural network with no hidden layer or nonlinearity: $\beta^{(m)}=g(\theta^{(m)};\omega)=\omega^{b}+\omega^{w}\theta^{(m)}$, where $\omega^{b} \in M(d_{x},1)$ and $\omega^{w} \in M(d_{x}, d_{\theta})$.
The predictions for time-series $m$ can hence be expressed as
\begin{equation}\label{eq:MtMsOLS}
    \hat{y}^{(m)}= X^{(m)}\underbrace{(\omega^{b}+\omega^{w}\theta^{(m)})}_{\beta^{(m)}}.
\end{equation}
This expression reveals that this special case of MtMs is analogous to performing PCA in the latent space of unobservable true regression coefficients $\{\beta^{*(m)}\}_{m=1}^{M}$.
The bias vector $\omega^{b}$ captures the central tendency of $\{\beta^{*(m)}\}_{m=1}^{M}$ and corresponds to the action of demeaning variables prior to PCA.
The column vectors of matrix $\omega^{w}$ are optimized to best explain the variability of the true unobserved $\{\beta^{*(m)}\}_{m=1}^{M}$, analogous to the loading vectors of individual principal components.
The task-specific parameter vector $\theta^{(m)}$ measures the exposure to variance-explaining factors $\omega^{w}[:,i]$ for a given time-series, and corresponds to the row $m$ of the score matrix from PCA.

Similarly to PCA, dimensionality reduction can be performed by choosing the number of factors ($d_{\theta}$) used to explain the variability of $\{\beta^{*(m)}\}_{m=1}^{M}$.
Choosing $d_{\theta}=0$ is equivalent to estimating a pooled regression on the time-series, whereas choosing $d_{\theta}$ = $d_{x}$ is equivalent to estimating a separate regression for each time-series $m$.
In practice, given that the DGPs of many time-series are likely similar, only a handful of factors $\omega^{w}[:,i]$ are necessary to successfully explain most of the variability across time-series.
Note that unlike principal component regression \citep[see, e.g.,][]{hadiCautionaryNotesUse1998}, where the dimensionality reduction is performed on the pooled design matrix as a preprocessing step, here the reduction is performed in the \emph{latent} space of regression coefficients jointly with the estimation.
In this sense, it is similar to reduced-rank regression \citep{izenmanReducedrankRegressionMultivariate1975}, with the exception that we are searching for a lower-dimensional representation of a set of regression coefficients across multiple tasks/time-series, rather than within a single dataset with multiple dependent variables.

Table \ref{tab:M4_losses_table} displays the average MASE for OLS localized via MtMs with $d_{\theta}=2$ on the M4 datasets.
To facilitate training, we leverage the fact that the optimal $\{\{\omega^{b}, \omega^{w}\}, \{\theta^{(m)}\}_{m=1}^{M}\}$ can be derived iteratively in closed form under the L2 loss\footnote{
    Eq. \ref{eq:MtMsOLS} can be expressed for all tasks $m$ simultaneously as  $\hat{\mathbf{y}}= \mathbf{X}(\tilde{\boldsymbol\theta} \,\otimes I_{d_{x}+1}) \textrm{vec}(\omega)$ where  $\hat{\mathbf{y}}=\left[\hat{y}^{(1)\top},\,\ldots,\,\hat{y}^{(M)\top}\right]^{\top}$, $\mathbf{X}=\textrm{blkdiag}(\{X^{(m)}\}_{m=1}^{M})$, $\omega = [\omega^{b}, \omega^{w}]$, $\boldsymbol\theta = \left[\theta^{(1)},\,\ldots,\,\theta^{(M)}\right]^{\top}$, and $\tilde{\boldsymbol\theta} = [\mathbf{1}, \boldsymbol\theta]$.
    This leads to first-order conditions for $\textrm{vec}(\omega)$: $\textrm{vec}(\omega)=\left(H^{\top}H\right)^{-1}H^{\top}\mathbf{y}$ where $\mathbf{y}=\left[y^{(1)\top},\,\ldots,\,y^{(M)\top}\right]^{\top}$ and $H=\mathbf{X}(\tilde{\boldsymbol\theta} \,\otimes I_{d_{x}+1})$.
    First-order conditions for $\{\theta^{(m)}\}_{m=1}^{M}$ are $\{\theta^{(m)}= \left(Q^{\top}Q\right)^{-1}Q^{\top}(y^{(m)}-X^{(m)}\omega^{b})\}_{m=1}^{M}$ where $Q = X^{(m)}\omega^{w}$.
    By iterating over these two sets of first-order conditions, $\{\{\omega^{b}, \omega^{w}\},\{\theta^{(m)}\}_{m=1}^{M}\}$ converge to their joint optimal values.
}, and use these estimates to initialize MtMs.
After initialization, the training is performed using backpropagation with the Adam optimizer, a learning rate of 0.001 and a minibatch size of 1000 time-series under the MASE loss.
For comparison with conventional localization techniques, we cluster time-series into $\{2^{i}\}_{i=2}^{10}$ clusters using k-means on \texttt{stl\_features} (seasonality \& trend), \texttt{entropy} and \texttt{acf\_features} (autocorrelation) from the \texttt{tsfeatures} package \citep{hyndmanTsfeaturesTimeSeries2023} and estimate regression coefficients for each cluster individually.
For each frequency, the number of lags $d_{x}$ is set to the maximum value according to the shortest series, following the setup of \citet{montero-mansoPrinciplesAlgorithmsForecasting2021}.
For reference, we also include the performance of OLS on the pooled dataset and two widely used local models: \texttt{ETS} \citep{hyndmanStateSpaceFramework2002a} and \texttt{auto.arima} \citep{hyndmanAutomaticTimeSeries2008a}.

With the exception of the yearly frequency, where localization provides only marginal improvements and where the two degrees of freedom per series likely lead to over-fitting, OLS localized via MtMs outperforms non-localized OLS and OLS localized via clustering across all cluster sizes.
Furthermore, for all frequencies except yearly, the simple linear parametric model derived in a data-driven way via MtMs, namely:
\begin{equation}
    \hat{y}_{t}^{(m)}=f_{\omega}(x_{t}^{(m)}; \theta^{(m)})=x_{t}^{(m)\top}(\omega^{b}+\omega^{w}\theta^{(m)})   
\end{equation}
with $\{\omega^{b}, \omega^{w}\}$ fixed, outperforms conventional and more complex local models crafted by human experts.

\begin{table}[!htbp]
    \fontsize{7}{7}\selectfont
    \centering
    \begin{filecontents*}{outputs/m4/losses_table.csv}
"new";"Yearly";"Quarterly";"Monthly";"Weekly"
"ETS";3.47843825340062;1.16430336776505;0.947684322375919;2.51272207198527
"auto.arima";3.40735349102804;1.1608330400368;0.92855415977213;2.54159074685201
"OLS (pooled)";3.05928026665811;1.22188007927296;0.956658107270366;2.2754626922943
"OLS (2 clusters)";3.01061091614019;1.21824393347308;0.950305391143852;2.24609342110075
"OLS (4 clusters)";2.99026645292999;1.22471149373823;0.95295752693821;2.17883288400968
"OLS (8 clusters)";3.01959401993187;1.22859847823142;0.949053732594917;2.19443004172752
"OLS (16 clusters)";3.07549433393094;1.22021823305341;0.951874641084071;2.18139990645929
"OLS (32 clusters)";3.13219469773422;1.21360783715178;0.950415108617823;
"OLS (64 clusters)";3.18791356370097;1.21013790891787;0.949371795815404;
"OLS (128 clusters)";3.25848085813588;1.20504181873488;0.942808689582233;
"OLS (256 clusters)";3.30377954540169;1.20128999283107;0.938573124480177;
"OLS (512 clusters)";3.40215127579056;1.19676797143473;0.93606595819537;
"OLS (1024 clusters)";3.53753350005166;1.19741851513992;0.93023446630821;
"OLS (localized via MtMs)";4.04964925646805;1.13330567250777;0.911246157356147;2.10400105530668
\end{filecontents*}
\pgfplotstabletypeset[
    col sep = semicolon,
    every row 4 column 1/.style={
        postproc cell content/.style={
        @cell content/.add={$\bf}{$}
        }
    },
    every row 13 column 2/.style={
        postproc cell content/.style={
        @cell content/.add={$\bf}{$}
        }
    },
    every row 13 column 3/.style={
        postproc cell content/.style={
        @cell content/.add={$\bf}{$}
        }
    },
    every row 13 column 4/.style={
        postproc cell content/.style={
        @cell content/.add={$\bf}{$}
        }
    },
    ignore chars={"},
    every head row/.style={before row={%
    \toprule
    },after row=\midrule},
    every last row/.style={after row=\bottomrule},
    display columns/0/.style={string type, column name={model}},
    display columns/1/.style={precision = 3, fixed, fixed zerofill=true, column name={Yearly}},
    display columns/2/.style={precision = 3, fixed, fixed zerofill=true, column name={Quarterly}},
    display columns/3/.style={precision = 3, fixed, fixed zerofill=true, column name={Monthly}},
    display columns/4/.style={precision = 3, fixed, fixed zerofill=true, column name={Weekly}}
    ]{outputs/m4/losses_table.csv}
    \caption{Losses for the M4 datasets\\
        Mean MASE losses for individual models on yearly, quarterly, monthly, and weekly datasets from the M4 competition.
        Bold text indicates the best-performing model for each frequency.
        Losses of OLS with more than $16$ clusters for the weekly frequency are not available due to an insufficient number of observations to estimate OLS on all clusters.
    }
    \label{tab:M4_losses_table}
\end{table}

\begin{figure}[!htbp]
    \centering
    \includegraphics[width=.99\linewidth]{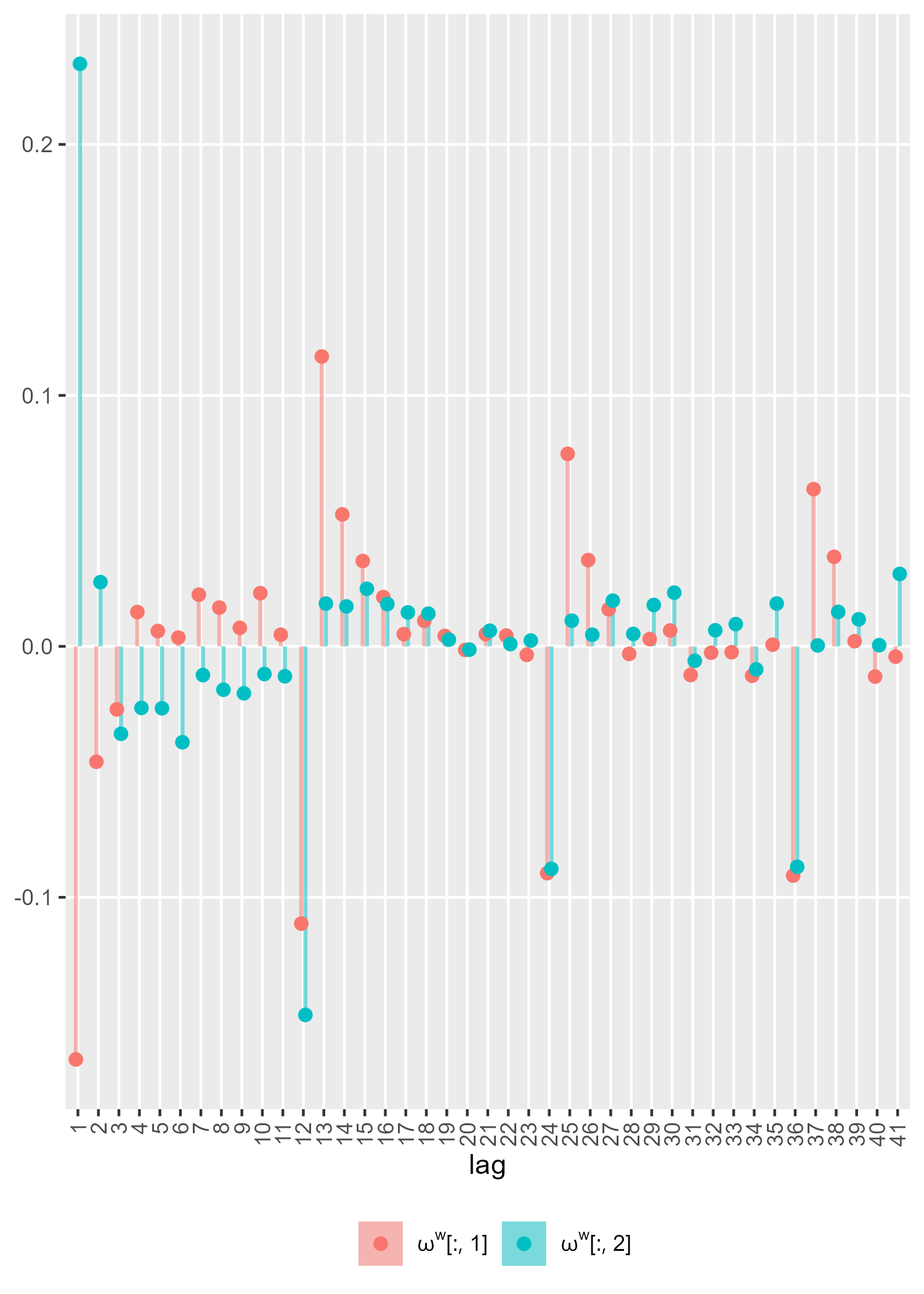}
    \caption{
        Estimated column vectors of $\omega^{w}$ for the M4 monthly dataset
    }
    \label{fig:m4_effects_monthly}
\end{figure}

Similarly to sinusoidal regression, the simple structure of the model allows us to visualize the heterogeneity across DGPs identified in the datasets.
As an example, Figure \ref{fig:m4_effects_monthly} displays the column vectors $\omega^{w}[:,1]$ and $\omega^{w}[:,2]$ for the monthly frequency.
Parameter $\theta[2]$ primarily regulates the persistence of the DGP (positively affecting the dependence on lag 1) and seasonality (negatively affecting the dependence on lags $\{12, 24, 36\}$).
To a lesser extent, it also captures seasonality at lag 6, likely driven by time-series with bi-annual seasonal behavior.
Parameter $\theta[1]$ also regulates persistence (negatively) and seasonality (negatively), but in addition appears to influence the decay of seasonal behavior, as evidenced by the gradually decreasing values of $\omega^{w}[:,1]$ corresponding to lags $\{13, 14, 25, 26, 37, 38\}$.
By varying these two parameters $\theta[1]$ and $\theta[2]$, one can approximately span the space of regression coefficients corresponding to DGPs encountered in the M4 monthly dataset.

\section{Supplementary materials}\label{section:supplemetary_materials}

\begin{table}[!htbp]
    \fontsize{7}{7}\selectfont
    \centering 
    \begin{tabular}{l l c} 
        \toprule
        Source & Feature                           & Transformation \\
        \midrule
        own    & Volatility(lag = [1,2,3,4,5,6,7]) &                \\
        own    & Return(lag = [1,2,3,4,5,6,7])     &                \\
        own    & IsETF                             &                \\
        TTR    & ADX                               &                \\
        TTR    & aroon                             &                \\
        TTR    & ATR(n=[7, 14, 28])                & Norm.          \\
        TTR    & BBands                            & Norm.          \\
        TTR    & CCI                               &                \\
        TTR    & chaikinAD                         & diff(1)        \\
        TTR    & chaikinVolatility                 &                \\
        TTR    & CLV                               &                \\
        TTR    & CMF                               &                \\
        TTR    & CMO                               &                \\
        TTR    & CTI                               &                \\
        TTR    & DEMA                              & Norm.          \\
        TTR    & DonchianChannel                   & Norm.          \\
        TTR    & EMA                               & Norm.          \\
        TTR    & EVWMA                             & Norm.          \\
        TTR    & GMMA(short=10, long=[30, 60])     & Norm.          \\
        TTR    & HMA                               & Norm.          \\
        TTR    & KST                               &                \\
        TTR    & MACD                              &                \\
        TTR    & MFI                               &                \\
        TTR    & OBV                               & diff(1)        \\
        TTR    & PBands                            & Norm.          \\
        TTR    & ROC                               &                \\
        TTR    & RSI                               &                \\
        TTR    & runPercentRank(n=100)             &                \\
        TTR    & SMI                               &                \\
        TTR    & SNR(n=[20,60])                    &                \\
        TTR    & TDI                               & Norm.          \\
        TTR    & TRIX                              &                \\
        TTR    & ultimateOscillator                &                \\
        TTR    & VHF                               &                \\
        TTR    & volatility                        &                \\
        TTR    & williamsAD                        & diff(1)        \\
        TTR    & WPR                               &                \\
        TTR    & ZLEMA                             & Norm.          \\
        \bottomrule
    \end{tabular}
    \caption{
        Features $x_{t}^{(m)}$ used as input to the model.
        The transformation ``Norm.'' indicates that the feature is normalized by the price of the asset while the transformation ``diff(1)'' denotes first differencing.
    } 
    \label{tab:features}
\end{table}

\begin{figure}[!htbp]
    \centering
    \includegraphics[width=1\linewidth]{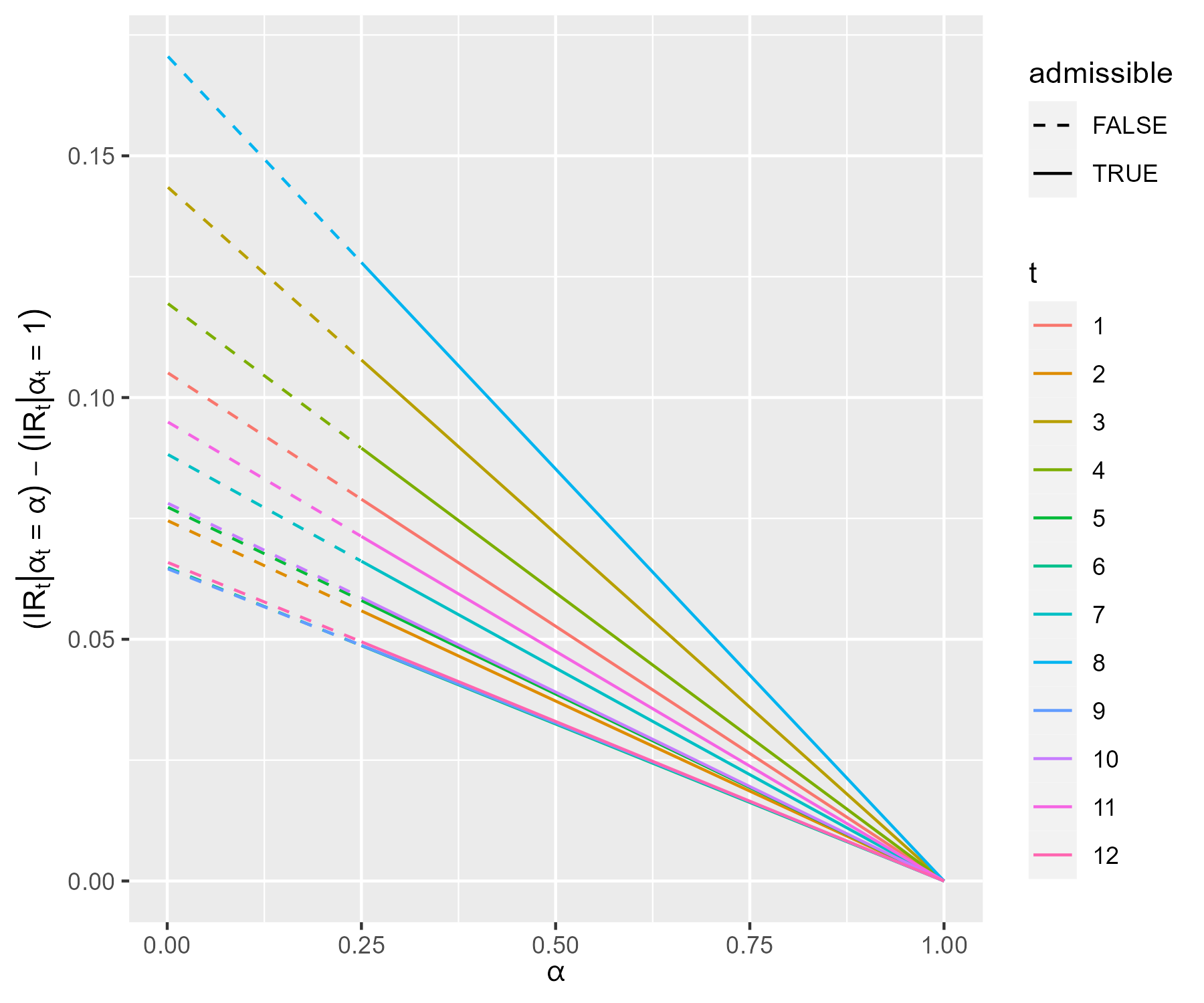}
    \caption{
        The difference in $IR_{t}$ for the long equal-weighted portfolio as a function of the scaling $\alpha_{t}$.
        Decreasing the scaling linearly improves $IR_{t}$.
    }
    \label{fig:scaling_effects}
\end{figure}

\section{Proofs}\label{section:proofs}

\begin{proof}
    Define $g(\theta^{(m)};\omega)=\{\theta^{(m)},\omega\}$ and $f(\cdot,\beta^{(m)})=f_{\beta^{(m)}[2]}(\cdot;\beta^{(m)}[1])$ with $B=\Theta \times \Omega$.
    Under A1 and A2, the bilevel optimization problem in Eq. \ref{eq:sample_bilevel_opt} reads as follows:
    \begin{equation}\label{eq:sample_bilevel_opt_alt}
        \begin{split}
            \hat{\omega}  =  \underset{\omega}{\arg\min} &\, \underbrace{\dfrac{1}{M} \sum_{m=1}^{M}   \dfrac{1}{K}\sum_{t=1}^{K} \gamma( y_{t}^{(m)}, f(x_{t}^{(m)};g(\theta^{(m)};\omega)))}_{\equiv Q(\omega,\{\kappa_{\omega}(\mathcal{D}_{train}^{(m)})\}_{m=1}^{M})}\\
            \textnormal{s.t.:} \,\hat{\theta}^{(m)} &= \kappa_{\omega} (\mathcal{D}_{train}^{(m)}) \\
            &= \underset{\theta}{\arg\min} \,  \dfrac{1}{K}\sum_{t=1}^{K} \gamma( y_{t}^{(m)}, f(x_{t}^{(m)};g(\theta;\omega))).\\
        \end{split}
    \end{equation}
    The assumption of existence and uniqueness of the inner optimization problems (A1) guarantees that the objective $Q(\omega,\{\kappa_{\omega}(\mathcal{D}_{train}^{(m)})\}_{m=1}^{M})$ is properly defined.
    Let us denote the set of solutions to the bilevel problem (Eq. \ref{eq:sample_bilevel_opt_alt}) as $\Omega_{B}^{*}\subset \Omega$, and the $\omega$ component of the set of solutions to the single-level problem (Eq. \ref{eq:sample_singlelevel_opt_alt}) as $\Omega_{S}^{*}$. 
    The fact that $ \Omega_{B}^{*} = \Omega_{S}^{*}$ directly stems from the fact that the individual components $(m)$ of the outer optimization objective $Q(\cdot)$ coincide with the inner optimization objectives:

    Let $\omega^{*} \in \Omega_{B}$.
    By virtue of optimality, $\forall \omega \in \Omega:\, Q(\omega^{*},\{\kappa_{\omega^{*}}(\mathcal{D}_{train}^{(m)})\}_{m=1}^{M}) \leq Q(\omega,\{\kappa_{\omega}(\mathcal{D}_{train}^{(m)})\}_{m=1}^{M})$.
    From the definition of $\kappa_{\omega}$ and the additivity of $Q(\cdot)$, it also holds $\forall \omega \in \Omega \, \forall \{\theta^{(m)}\}_{m=1}^{M} \in \Theta^{M}:\, Q(\omega,\{\kappa_{\omega}(\mathcal{D}_{train}^{(m)})\}_{m=1}^{M}) \leq Q(\omega,\{\theta^{(m)}\}_{m=1}^{M})$.
    Combining these, we obtain $\forall \omega \in \Omega \, \forall \{\theta^{(m)}\}_{m=1}^{M} \in \Theta^{M}:\,Q(\omega^{*},\{\kappa_{\omega^{*}}(\mathcal{D}_{train}^{(m)})\}_{m=1}^{M}) \leq Q(\omega,\{\theta^{(m)}\}_{m=1}^{M})$ which implies $\omega^{*} \in \Omega_{S}$.

    Let $\omega^{*} \in \Omega_{S}$, and let $\{\theta^{*(m)}\}_{m=1}^{M} \in \Theta^{M}$ be the corresponding $\theta$ component of the solution.
    By virtue of optimality, $\forall \omega \in \Omega \, \forall \{\theta^{(m)}\}_{m=1}^{M} \in \Theta^{M}:\,Q(\omega^{*},\{\theta^{*(m)}\}_{m=1}^{M}) \leq Q(\omega,\{\theta^{(m)}\}_{m=1}^{M})$.
    Since $\kappa_{\omega}(\mathcal{D}_{train}^{(m)}) \in \Theta$, it follows that $\forall \omega \in \Omega:\,Q(\omega^{*},\{\theta^{*(m)}\}_{m=1}^{M}) \leq Q(\omega,\{\kappa_{\omega}(\mathcal{D}_{train}^{(m)})\}_{m=1}^{M})$.
    From the definition of $\kappa_{\omega}$ and additivity of $Q(\cdot)$, it also holds $Q(\omega^{*},\{\kappa_{\omega^{*}}(\mathcal{D}_{train}^{(m)})\}_{m=1}^{M}) \leq Q(\omega^{*},\{\theta^{*(m)}\}_{m=1}^{M})$.
    Combining these, we obtain $\forall \omega \in \Omega:Q(\omega^{*},\{\kappa_{\omega^{*}}(\mathcal{D}_{train}^{(m)})\}_{m=1}^{M}) \leq Q(\omega,\{\kappa_{\omega}(\mathcal{D}_{train}^{(m)})\}_{m=1}^{M})$, which implies $\omega^{*} \in \Omega_{B}$.
\end{proof}

\bibliographystyle{elsarticle-harv}\biboptions{authoryear}
\bibliography{Library.bib}

\end{document}